\documentclass{bmvc2k}

\title{Label Smoothing++: Enhanced Label Regularization for Training Neural Networks}

\addauthor{Sachin Chhabra}{sachin.chhabra@asu.edu}{1}
\addauthor{Hemanth Venkateswara}{hvenkateswara@gsu.edu}{2}
\addauthor{Baoxin Li}{baoxin.li@asu.edu}{1}

\addinstitution{
 Arizona State University\\
 699 S Mill Ave.\\
 Tempe, AZ, USA- 85281
}
\addinstitution{
 Georgia State University\\
 25 Park Pl NE\\
 Atlanta, GA, USA - 30303
}

\runninghead{Chhabra, Venkateswara, Li}{Label Smoothing++}

\usepackage{booktabs}
\usepackage{array} 
\usepackage{nicematrix}

\definecolor{lightgray}{gray}{0.95}

\begin{document}

\maketitle

\begin{abstract}
Training neural networks with one-hot target labels often results in overconfidence and overfitting. Label smoothing addresses this issue by perturbing the one-hot target labels by adding a uniform probability vector to create a regularized label. 
Although label smoothing improves the network's generalization ability, it assigns equal importance to all the non-target classes, which destroys the inter-class relationships. 
In this paper, we propose a novel label regularization training strategy called Label Smoothing++, which assigns non-zero probabilities to non-target classes and accounts for their inter-class relationships.
Our approach uses a fixed label for the target class while enabling the network to learn the labels associated with non-target classes.
% Our approach assigns non-zero probabilities to non-target classes and accounts for their inter-class relationships.
Through extensive experiments on multiple datasets, we demonstrate how Label Smoothing++ mitigates overconfident predictions while promoting inter-class relationships and generalization capabilities. 
\end{abstract}

\section{Introduction}
One of the most common practices for training neural networks is cross-entropy loss with one-hot target labels. 
However, it has been shown that this leads to overfitting and overconfident predictions by the network \cite{ls}. 
Numerous regularization techniques have been proposed to impose additional constraints to tackle this issue. 
Some of these techniques like Cutout \cite{cutout}, Mixup \cite{mixup}, CutMix \cite{cutmix}, and others \cite{augmix,keepaugment,patchswap} alter the input data and are applied without considering the object positions, potentially impacting such entities directly. 
An alternative approach is label regularization, which operates on training labels.
Label smoothing is one of the easiest methods that create regularized targets by taking a weighted sum of the one-hot probability vector and a uniform vector based on a hyperparameter $\alpha$ to mitigate the overconfidence problem.  

Nowadays, Label Smoothing has become one of the standard ways of training neural networks \cite{shot,gap}. 
Even though it provides benefits in the form of generalization, it is known to eliminate inter-class relationships \cite{lshelp}. 
By using a uniform probability vector, Label Smoothing assigns equal weight to all the non-target classes, which means all the classes are equally different from the target class. 
However, this is not always the case. 
For example, consider a 4-way classification of \textit{Bird}, \textit{Car}, \textit{Frog}, and \textit{Truck}. 
Here, the training label for the \textit{Car} class should have more weight assigned to \textit{Truck} given their semantic similarity as compared to other classes like \textit{Frog} and \textit{Bird}.
Such inter-class relationships, overlooked by Label Smoothing, can help enhance generalization abilities, knowledge distillation, learning from noisy labels, and handling missing data \cite{distill,lshelp,ols}. 

\begin{figure*}[t]
\centering
     \begin{subfigure}[b]{0.18\textwidth}
         \centering
         \includegraphics[width=\textwidth]{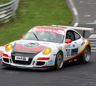}
         \caption{Input}
     \end{subfigure}
     \hfill
     \begin{subfigure}[b]{0.25\textwidth}
         \centering
         \includegraphics[width=\textwidth]{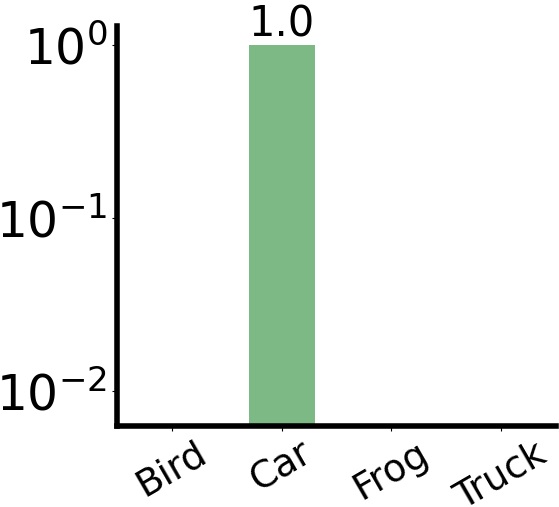}
         \caption{Vanilla}
     \end{subfigure}
     \hfill
     \begin{subfigure}[b]{0.2\textwidth}
         \centering
         \includegraphics[width=\textwidth]{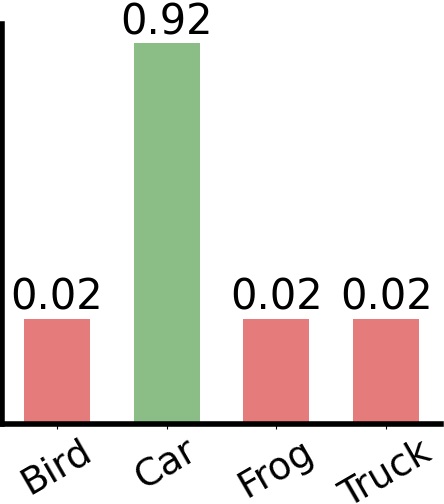}
         \caption{LS}
     \end{subfigure}
     \hfill
     \begin{subfigure}[b]{0.2\textwidth}
         \centering
         \includegraphics[width=\textwidth]{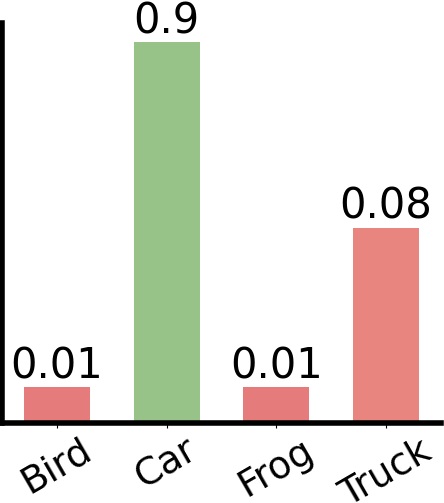}
         \caption{LS\texttt{+}\texttt{+}}
     \end{subfigure}
    \caption{Different types of labels generated by various sources. 
    (a) Input.
    (b) Traditionally (Vanilla), 1-hot probability vectors are used as labels.
    (c) Label Smoothing (LS) evenly distributes a probability parameter (denoted as $\alpha$, with a value of 0.1 in this context) across all classes.
    (d) Label Smoothing (LS$\texttt{++}$) distributes $\alpha$ probability among the non-target classes for all classes independently.}
    \label{fig:toy_probs}
\end{figure*}

This paper proposes a novel label regularization method called Label Smoothing$\texttt{++}$ (LS$\texttt{++}$) that generates regularized training labels from one-hot target labels. 
It is done by retaining the confidence of the target class to be high while also assigning non-zero probabilities to the non-target classes by accounting for inter-class relationships. 
In standard Label Smoothing, the 1-hot target label is perturbed by adding a uniform distribution to the 1-hot target vector. 
In Label Smoothing$\texttt{++}$, we determine a class-wise probability vector to add to the 1-hot vector. 
Here, the samples of a class are constrained to produce the same outputs for all the classes. 

% offers multiple advantages while training neural networks. 
Label Smoothing$\texttt{++}$  provides flexibility in how the probabilities are assigned among the non-target classes, which is essential for learning inter-class relationships. 
Refer to Figure \ref{fig:toy_probs} where we display targets generated by different label regularization techniques for a 4-class classification problem. 
Through experiments on multiple datasets and in various settings, we show the strengths of our proposed method, Label Smoothing$\texttt{++}$, compared to other label regularization techniques.

\section{Related Work}
Using a 1-hot target label in neural network training is known to lead to overconfidence and hinder generalization \cite{ls}. 
% Various regularization techniques aim to address this issue, primarily focusing on modifying input data \cite{mixup,cutmix,cutout}. 
Label regularization techniques aim to modify training labels to mitigate overconfidence. 
One of the earliest and easiest label regularization methods is Label Smoothing, which combines the 1-hot vector with a uniform vector based on a hyperparameter $\alpha$ \cite{ls}. 
Despite its advantages, Label Smoothing destroys inter-class relationships by assigning equal weights to all non-target classes \cite{lshelp}. 
We aim to deviate from using a uniform vector by offering the network the flexibility to adjust its training label.

An alternative for regularizing network predictions is entropy maximization \cite{entmax}. 
Entropy maximization directly penalizes the network for overconfident predictions.
This technique provides greater flexibility but requires hyperparameter tuning for the entropy maximization loss weight.
Focal loss is a modification of the cross-entropy loss function that was introduced to address overconfidence by assigning higher weights to samples with low confidence and lower weights to those with high confidence \cite{focal,focalorg}. 
This approach minimizes entropy maximization and a regularized KL divergence to prevent the network from becoming excessively overconfident.

Knowledge distillation is considered a form of label regularization that involves generating targets from a larger network (the Teacher) and transferring this knowledge to a smaller network (the Student) on a per-sample basis \cite{distill,tfkd}. 
The relationship of each sample to non-target classes, as learned by the Teacher, helps regulate the student networks \cite{distill}. 
In alignment with this concept, a trained network was used to train another network with the same architecture in Teacher-Free Knowledge Distillation \cite{tfkd}. 
However, this approach incurs significant computational expenses as it requires training a network twice and generating outputs online. 
An alternative, Teacher-Free regularization, behaves similarly to Label Smoothing but utilizes a high mixing coefficient of $0.9$ to generate a smoothed probability vector \cite{tfkd}. 
The network is trained to align predicted probabilities with this vector at a high temperature, reducing computational costs but still relying on a uniform vector. 

Online Label Smoothing is another label regularization approach that is based on network predictions \cite{ols}. 
It computes average network predictions for each class and mixes them with a 1-hot probability vector.
While it diminishes the need to train the network twice, it carries a substantial computational overhead as average network predictions must be computed every epoch on the training set.
Our approach also has a class-based alignment (without the computational overhead of computing it every epoch) but only allows changes in the distribution of probabilities among the non-target classes, unlike online Label Smoothing, where training labels become 1-hot when network predictions tend towards 1-hot.

\section{Methodology}
\subsection{Background}
Consider a dataset $D:=\{(x_i, y_i)\}_{i=1}^m$ with $K$ classes. For a pair $(x,y)$, $x$ is the input and $y$ is its corresponding target with $y \in \{1,2,\ldots,K\}$. 
Let $\bar{y}=[\bar{y}_1,\bar{y}_2,\ldots,\bar{y}_K]^\top$ denote the one-hot presentation of the target label $y$. 
A neural network $G$ takes $x$ as input and generates a probability vector $G(x)=\hat{y}=[\hat{y}_1,\hat{y}_2, \dots, \hat{y}_K]$. 
Here, $\hat{y}$ is the predicted probability for the input $x$. 
The traditional procedure for training $G$ is minimizing the cross-entropy loss,
\begin{flalign}
    H(\bar{y},\hat{y}) = - \bar{y} \log \hat{y} = -\sum_{i=1}^{K} \bar{y}_i \log \hat{y}_i = -\log \hat{y}_y.
\end{flalign}
Training a neural network with 1-hot target labels often leads to issues of overconfidence and overfitting \cite{ls}. Label smoothing is a popular label regularization technique that alleviates this problem.
It modifies the training label by using a weighted combination of the 1-hot and a uniform probability vector.
\begin{equation}
\bar{y}^{ls} = (1-\alpha)\bar{y} + \alpha u,
\label{Eq:ls}
\end{equation}
where $u=[\frac{1}{K},\ldots,\frac{1}{K}]^\top$ is a uniform probability vector of size $K$ and each element is equal to $\frac{1}{K}$.
$\alpha$ is a hyperparameter that decides the weight between 1-hot and the uniform probability vector.
Label Smoothing trains the network minimizing the same cross-entropy loss but with regularized training label: 
$H(\bar{y}^{ls},\hat{y}) = - \bar{y}^{ls} \log \hat{y} = -\sum_{i=1}^{K} \bar{y}^{ls}_i \log \hat{y}_i$.
% \begin{equation}
% H(\bar{y}^{ls},\hat{y}) = - \bar{y}^{ls} \log \hat{y} = -\sum_{i=1}^{K} \bar{y}^{ls}_i \log \hat{y}_i
% \end{equation}
% \begin{flalign}
%     H(\bar{y}^{ls},\hat{y}) &= - \bar{y}^{ls} \log \hat{y} = -\sum_{i=1}^{K} \bar{y}^{ls}_i \log \hat{y}_i \\
%     &=  -\sum_{i=1}^{K} [(1-\alpha)\bar{y} + \alpha u] \log \hat{y}_i \notag \\
%     &=  -(1-\alpha)\log \hat{y}_y - \alpha \sum_{i=1}^{K} u \log \hat{y}_i \notag \\
%     &=  (1-\alpha)H(\bar{y},\hat{y}) + \alpha H(u,\hat{y}) \notag 
% \end{flalign}
% Training with Label Smoothing is essentially a weighted combination of cross-entropy loss of predictions with one-hot targets and uniform vectors.

\begin{figure*}[t]
\centering
    \includegraphics[width=\textwidth]{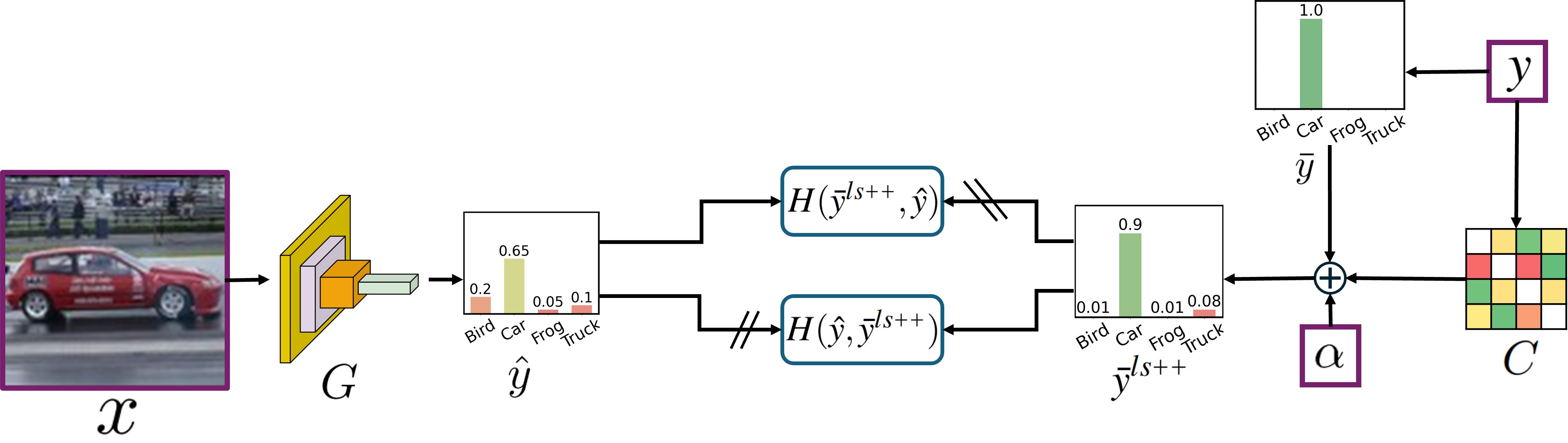}
    \caption{Model Diagram of Label Smoothing$\texttt{++}$ (LS$\texttt{++}$). Our approach distributes $\alpha$ confidence among the non-target classes using learnable targets for each class independently. This promotes all samples of a class to achieve similar output. We train Label Smoothing$\texttt{++}$ with symmetric cross-entropy loss but because of learnable targets, we stop the flow of gradients from loss to some parameters (visually represented by double-slant lines).}
    \label{fig:modeldig}
\end{figure*}

\subsection{Label Smoothing$\texttt{++}$}
\label{sec:lspp}
The idea behind label regularization is to reduce the confidence of the target class by a value $\alpha$ and increment the confidence of all classes by the same amount $\alpha$. 
Label Smoothing distributes the value $\alpha$ uniformly among all the classes.
Instead of using a uniform assignment, we propose to learn the optimal assignment.
We assume that the samples of a class generally share some similar characteristics. So, their output vector should share similarities as well.
With this understanding, we propose Label Smoothing$\texttt{++}$ (LS$\texttt{++}$). 

We train the network to learn the assignment of the residual probability $\alpha$ among the non-target classes for each target class independently. 
% based on the similarity of the target class with the non-target classes. 
For every class $y$ we learn a probability vector $C_y$ of length $(K-1)$ which represents the probability assignment of the non-target $K-1$ classes. 
We used a $K-1$ length to distribute the residual $\alpha$ probability among the non-target classes only.
This ensures the probability of the target class remains unchanged and only the non-target classes are adjusted.
The regularized training label is then given by,  
\begin{equation}
\bar{y}^{ls\texttt{+}\texttt{+}} = (1-\alpha)\bar{y} + \alpha C_y.
\label{ylspp_eq}
\end{equation}
Note: $C_y$ has size $K-1$ but is adjusted to be length $K$ after inserting a $0$ at the ground truth position $y$ in eq. \ref{ylspp_eq}. 
% \begin{equation}
%     \bar{y}^{ls\texttt{+}\texttt{+}} = 
% \begin{cases}
%     1-\alpha & \text{if } i=y,\\
%     \alpha C_{yi}       & \text{otherwise}
% \end{cases}
% \
% \end{equation}
We train the network to predict the same regularized training label for all samples belonging to a class.  
The $\{C_y\}_{y=1}^K$ vectors together form the $C$ matrix of dimension $K \times K$ which is estimated by training. 
The $C$-Matrix has a diagonal element set to 0.
Label Smoothing$\texttt{++}$ provides the same mixing training label for samples belonging to a class but different for each class.
Our goal here is to provide freedom to the network to choose its optimal training label while adhering to the class-level constraint. 
Since the label for the target class is fixed, we only need to impose consistency on the non-target classes. Note: The $C$-Matrix is not a symmetrical matrix, as class relationships can differ based on the query class. We discuss this in more detail in the supplementary material.

Label Smoothing$\texttt{++}$ has training labels $\bar{y}^{ls\texttt{+}\texttt{+}}$ that need to be learned. 
For training the network, we depart from the traditional cross-entropy loss, which works well only for fixed training labels. 
The cross-entropy loss is an upper-bound on the Kullback-Leibler (KL) divergence with, $H(\bar{y}^{ls\texttt{+}\texttt{+}},\hat{y}) = KL(\bar{y}^{ls\texttt{+}\texttt{+}}||\hat{y}) + H(\bar{y}^{ls\texttt{+}\texttt{+}})$, where $H(\bar{y}^{ls\texttt{+}\texttt{+}})$ is the entropy of the training label $\bar{y}^{ls\texttt{+}\texttt{+}}$. 
When the training label is fixed, such as in 1-hot or Label Smoothing, the entropy is merely a constant. 
But with $\bar{y}^{ls\texttt{+}\texttt{+}}$, the training label contains learnable parameter $C$. 
Applying an entropy minimization loss on $\bar{y}^{ls\texttt{+}\texttt{+}}$ results in assigning all the probability to one of the classes in $C_y$, which is undesirable. 
We address this using a symmetric cross-entropy loss $H(\bar{y}^{ls\texttt{+}\texttt{+}}, \hat{y})+H(\hat{y}, \bar{y}^{ls\texttt{+}\texttt{+}})$ where the network parameters $G$ are trained using the first term $H(\bar{y}^{ls\texttt{+}\texttt{+}}, \hat{y})$, and the $C$ matrix is trained using the second term $H(\hat{y}, \bar{y}^{ls\texttt{+}\texttt{+}})$. 
The $H(\bar{y}^{ls\texttt{+}\texttt{+}},\hat{y})$ loss updates only the parameters in $G$ and does not affect $C$ thereby negating the effect of the entropy term $H(\bar{y}^{ls\texttt{+}\texttt{+}})$. Likewise, $H(\hat{y}, \bar{y}^{ls\texttt{+}\texttt{+}})$ loss updates only matrix $C$. 
%is trained using  we allow gradients of each loss only for the second term in the loss function. 
%To be specific $H(\bar{y}^{ls\texttt{+}\texttt{+}}, \hat{y})$ is used to train the network $G$ only and the loss $H(\hat{y}, \bar{y}^{ls\texttt{+}\texttt{+}})$ is used to train only the matrix $C$.
%This nullifies the effect of entropy minimization on all the parameters and only KL divergence is reduced between the targets and the predictions.

\begin{table*}[t]
    % \small
    \footnotesize
    \begin{center}
    \begin{NiceTabular}{|l|ccccc|}
        \toprule
        Method  & ResNet18  & ResNet34  & ResNet50  & ResNet101 &  DenseNet121\\
        \midrule
         1-hot & 75.87 & 79.38 & 78.79 & 79.66 & 79.04\\
        LS  & 77.26 & 79.06 & 78.80 & 79.88 & 80.38\\
        TFKD$_{self}$ & 77.10 & - & - & -  & 80.26\\
        TFKD$_{reg}$ & 77.36 & - & - & -  & -\\
        OLS  & - & 79.96 & 79.35 & 80.34& -\\
        % Zipf's LS \cite{zipf} & 77.38$\pm$0.32 & & & & 79.03$\pm$0.32\\ 
        Zipf's LS & 77.38 & & & & 79.03\\ 
        FL-3  & -& -&77.25& -& -\\
        FLSD-53 & - & -&76.78&  -& -\\
        \midrule
        LS$\texttt{++}$ (Ours) & \textbf{79.33$\pm$0.23} & \textbf{80.25$\pm$0.14} & \textbf{81.05$\pm$0.73} & \textbf{81.13$\pm$0.52} & \textbf{80.71$\pm$0.13}\\
        \bottomrule
    \end{NiceTabular}
    \end{center}
    \caption{Top-1 Classification accuracy on CIFAR100 dataset using different networks.}
    \label{tab:c100_results}
\end{table*}

% \begin{table*}[t]
%     % \small
%     \footnotesize
%     \begin{center}
%     \begin{NiceTabular}{|l|ccccc|}
%         \toprule
%         Method  & ResNet18  & ResNet34  & ResNet50  & ResNet101 &  DenseNet121\\
%         \midrule
%          1-hot & 75.87 & 79.38 & 78.79 & 79.66 & 79.04\\
%         LS \cite{ls} & 77.26 & 79.06 & 78.80 & 79.88 & 80.38\\
%         TFKD$_{self}$ \cite{tfkd} & 77.10 & - & - & -  & 80.26\\
%         TFKD$_{reg}$ \cite{tfkd} & 77.36 & - & - & -  & -\\
%         OLS \cite{ols} & - & 79.96 & 79.35 & 80.34& -\\
%         % Zipf's LS \cite{zipf} & 77.38$\pm$0.32 & & & & 79.03$\pm$0.32\\ 
%         Zipf's LS \cite{zipf} & 77.38 & & & & 79.03\\ 
%         FL-3 \cite{focal} & -& -&77.25& -& -\\
%         FLSD-53 \cite{focal} & - & -&76.78&  -& -\\
%         \midrule
%         LS$\texttt{++}$ (Ours) & \textbf{79.33$\pm$0.23} & \textbf{80.25$\pm$0.14} & \textbf{81.05$\pm$0.73} & \textbf{81.13$\pm$0.52} & \textbf{80.71$\pm$0.13}\\
%         \bottomrule
%     \end{NiceTabular}
%     \end{center}
%     \caption{Top-1 Classification accuracy on CIFAR100 dataset using different networks.}
%     \label{tab:c100_results}
% \end{table*}

\section{Experiments}
\subsection{Datasets and Setup}
We conducted extensive testing of our approach across a range of datasets, including FashionMNIST \cite{fmnist}, SVHN \cite{svhn}, CIFAR10 \cite{cifar}, CIFAR100 \cite{cifar}, FER2013, Animals10N \cite{animals}, Tiny-ImageNet, and ImageNet-100, employing various network architectures \cite{resnet,shufflenet,densenet,lenet,alexnet,vit}.
% (Results on Vision Transformers are in the supplementary material).
% These datasets offer a diverse array of variations for comprehensive testing.
Due to hardware limitations, we used Tiny-ImageNet and ImageNet-100 as substitutes for the original ImageNet dataset \cite{imagenet}. 
Tiny-ImageNet features $64\times 64$ images with 200 classes, while ImageNet-100 uses 100 classes and the original $224\times 224$ image size.

Our methodology was also applied to non-image modalities like Video, Text, and Audio. 
In the case of the video modality, we utilized UCF101 \cite{ucf101} and HMDB51 \cite{hmdb51} datasets, employing Conv+LSTM (CLSTM) and a C3D \cite{c3d} networks. 
The Conv+LSTM network utilized a ResNet50 pre-trained on ImageNet as the backbone, with the LSTM layers trained from scratch. 
The C3D network is a 3D convolution network pre-trained on the Sports-1M dataset \cite{sports1m}. 
For the text modality, our approach was tested on 20Newsgroup, AGNews \cite{agnews}, and YahooAnswers \cite{agnews} datasets using pre-trained BERT model \cite{bert}.
For the Audio modality, we used MelSpectrograms of the GTZAN \cite{gtzan} and SpeechCommands \cite{speechcommands} datasets. 
CNN models pre-trained on ImageNet have shown enhanced generalization on the audio domain \cite{DBLP:journals/corr/abs-2007-11154}. 
Hence, we trained ResNet50 from scratch and also tested a pre-trained network. 
Full details of the training augmentations and other details for all the datasets is available in the supplementary material. 
% We assessed our methodology by training ResNet50 from scratch and using a pre-trained ResNet50. 

\begin{table*}[t]
    % \small
    \footnotesize
    \begin{center}
    \begin{NiceTabular}{|l|ccccc|}
        \toprule
        Method  & ResNet18  & ResNet50  & ResNet101  & ShuffleNet &  DenseNet121\\
        \midrule
        1-hot & 64.33 & 67.47 & 69.03 & 60.51 & 68.15\\
        LS & 64.74 & 67.63 & 69.30 & 60.66 & 68.19\\
        TFKD$_{self}$ & - & 68.18 & - & 61.36 & 68.29\\
        TFKD$_{reg}$ & - & 68.15 & - & 60.93 & 68.37\\
        MBLS & - & 65.15 & 65.81 & - & -\\
        % Zipf's LS  & 59.25$\pm$0.20 & & & &  62.64$\pm$0.30\\
        Zipf's LS & 59.25 & & & &  62.64\\
        FL-3 & - & 50.31 & 62.97 & - & -\\
        FLSD-53 & - & 50.94 & 62.96 & - & -\\
        \midrule
        LS$\texttt{++}$ (Ours) & \textbf{65.07$\pm$0.08} & \textbf{69.01$\pm$0.46} & \textbf{70.04$\pm$0.30} & \textbf{63.24$\pm$0.49} & \textbf{68.90$\pm$0.11}\\
        \bottomrule
    \end{NiceTabular}
    \end{center}
    \caption{Top-1 Classification accuracy on Tiny-ImageNet dataset using different networks.}
    \label{tab:ti_results}
\end{table*}

\begin{table*}[!t]
    \footnotesize
    % \small
    \begin{center}
    % \begin{NiceTabular}{|l|c|c|c|c|c|cc|}
    \begin{NiceTabular}{|p{1.1cm}|
    >{\centering\arraybackslash}p{1.3cm} | 
    >{\centering\arraybackslash}p{1.2cm} | >{\centering\arraybackslash}p{1.3cm} | 
    >{\centering\arraybackslash}p{1.2cm} | >{\centering\arraybackslash}p{1.3cm} |
    >{\centering\arraybackslash}p{0.8cm} >{\centering\arraybackslash}p{1cm} |
}
        \toprule
        Dataset & FMNIST & SVHN & CIFAR10 & FER & Animals10N & \multicolumn{2}{c|}{ImageNet-100}\\
        \midrule
        Network & LeNet & LeNet & AlexNet & ResNet18 & ResNet18 & ResNet18 & ResNet50\\
        \midrule
        1hot & 82.23$\pm$0.34 & 89.40$\pm$0.03 & 79.98$\pm$0.17  & 70.10$\pm$0.21
& 85.00$\pm$0.11 & 81.72 &  83.96\\
        LS & 82.55$\pm$0.62 & 89.35$\pm$0.09 & 80.66$\pm$0.20 &  70.61$\pm$0.10
& 86.13$\pm$0.19 & 82.22 & 84.58\\
        TFKD$_{reg}$ & 82.40$\pm$0.26 & 89.42$\pm$0.31 &  80.78$\pm$0.17 & \textbf{70.80$\pm$0.41}
& 85.99$\pm$0.10 & 82.44 & 84.72\\
        OLS & 82.97$\pm$0.50 &  89.19$\pm$0.43 & 80.71$\pm$0.28 & 70.67$\pm$0.17
 & 86.35$\pm$0.38 & 82.56 & 84.71\\ 
        % \midrule
        LS++&  \textbf{83.79$\pm$0.23} &  \textbf{89.77$\pm$0.21} & \textbf{81.19$\pm$0.05} & \textbf{70.80$\pm$0.22} & \textbf{86.51$\pm$0.15} & \textbf{82.70} & \textbf{85.06}\\
        \bottomrule
    \end{NiceTabular}
    \end{center}
    \caption{Top-1 Classification accuracy on FashionMNIST (FMNIST), SVHN, CIFAR10, Facial expression recognition (FER), Animals10N, and ImageNet-100 datasets.}
    \label{tab:im100_results}
\end{table*}

\begin{table*}[!t]
    \footnotesize
    \begin{center}        
    % \begin{NiceTabular}{|l|cc|cc|ccc|cc|cc|}
    \begin{NiceTabular}{|p{1.1cm}|
    >{\centering\arraybackslash}p{0.7cm}  >{\centering\arraybackslash}p{0.7cm}|
    >{\centering\arraybackslash}p{0.6cm}  >{\centering\arraybackslash}p{0.8cm}|
    >{\centering\arraybackslash}p{0.4cm}  >{\centering\arraybackslash}p{0.6cm}   >{\centering\arraybackslash}p{0.5cm} |
    >{\centering\arraybackslash}p{0.5cm} >{\centering\arraybackslash}p{0.5cm}|
    >{\centering\arraybackslash}p{0.5cm} >{\centering\arraybackslash}p{0.5cm}|}
        \toprule
        Modality & \multicolumn{4}{c|}{Video} & \multicolumn{3}{c|}{Text} & \multicolumn{4}{c|}{Audio}\\
        \midrule
        Dataset & \multicolumn{2}{c|}{UCF101} & \multicolumn{2}{c|}{HDMB51} & 20NG & AGNews & YA & \multicolumn{2}{c|}{GTZAN} & \multicolumn{2}{c|}{SC}\\
        \midrule
        Network  & CLSTM$^*$ & C3D$^*$ & CLSTM$^*$ & C3D$^*$ & \multicolumn{3}{c|}{BERT$^*$} & R50* & R50 & R50* & R50\\
        \midrule
        1-hot & 71.13 & 78.56 & 36.01 & 45.88 & 85.02 & 94.39 & 77.44 & 91.50 & 87.50 & 96.03 & 95.13\\
        LS & 71.87 & 81.82 & 37.91 & 49.74 & 85.15 & 94.50 & 77.51 & 92.50 & 87.50 & \textbf{96.22} & 95.29\\
        LS$\texttt{++}$ & \textbf{72.56} & \textbf{82.37} & \textbf{38.24} & \textbf{51.31} & \textbf{85.55} & \textbf{94.67} & \textbf{77.54} & \textbf{93.50} & \textbf{89.00} & \textbf{96.22} & \textbf{95.45}\\
        \bottomrule
    \end{NiceTabular}
    \end{center}
    \caption{Comparison of Top-1 test accuracies on Video, Text, and Audio Modalities. CLSTM: Convolution + LSTM. $*$ denotes a pre-trained network was finetuned.}
    \label{tab:modal_results}
\end{table*}

Across all tasks, we trained Label Smoothing$\texttt{++}$ with a consistent setting of $\alpha=0.1$.
The matrix $C$ is stored as pre-softmax values (logits) and was initialized with zeros, resulting in a uniform probability distribution for the non-target classes as the starting point.
The code for Label Smoothing$\texttt{++}$ can be found at \url{https://github.com/s-chh/LSPP}.
% The code and the pseudo code are also available in the supplementary material
% and will be made public upon acceptance.

\subsection{Results}
We conducted a comprehensive evaluation of Label Smoothing$\texttt{++}$ (LS$\texttt{++}$) against other label regularization techniques, including Label Smoothing (LS) \cite{ls}, Online Label Smoothing (OLS) \cite{ols}, Margin-based Label Smoothing (MBLS) \cite{mbls}, Teacher-Free Knowledge Distillation (TFKD) \cite{tfkd}, and Focal loss \cite{focal,focalorg}. The summarized results can be found in Tables \ref{tab:c100_results}, \ref{tab:ti_results}, \ref{tab:im100_results}, and \ref{tab:modal_results}.
`-' indicates results were not available in the original paper.
Notably, for Tables \ref{tab:im100_results} and \ref{tab:modal_results}, baseline experiments were conducted by us using the same setup as ours.

\begin{figure*}[!t]
\centering
     \begin{subfigure}[b]{0.27\textwidth}
         \centering
         \includegraphics[width=\textwidth]{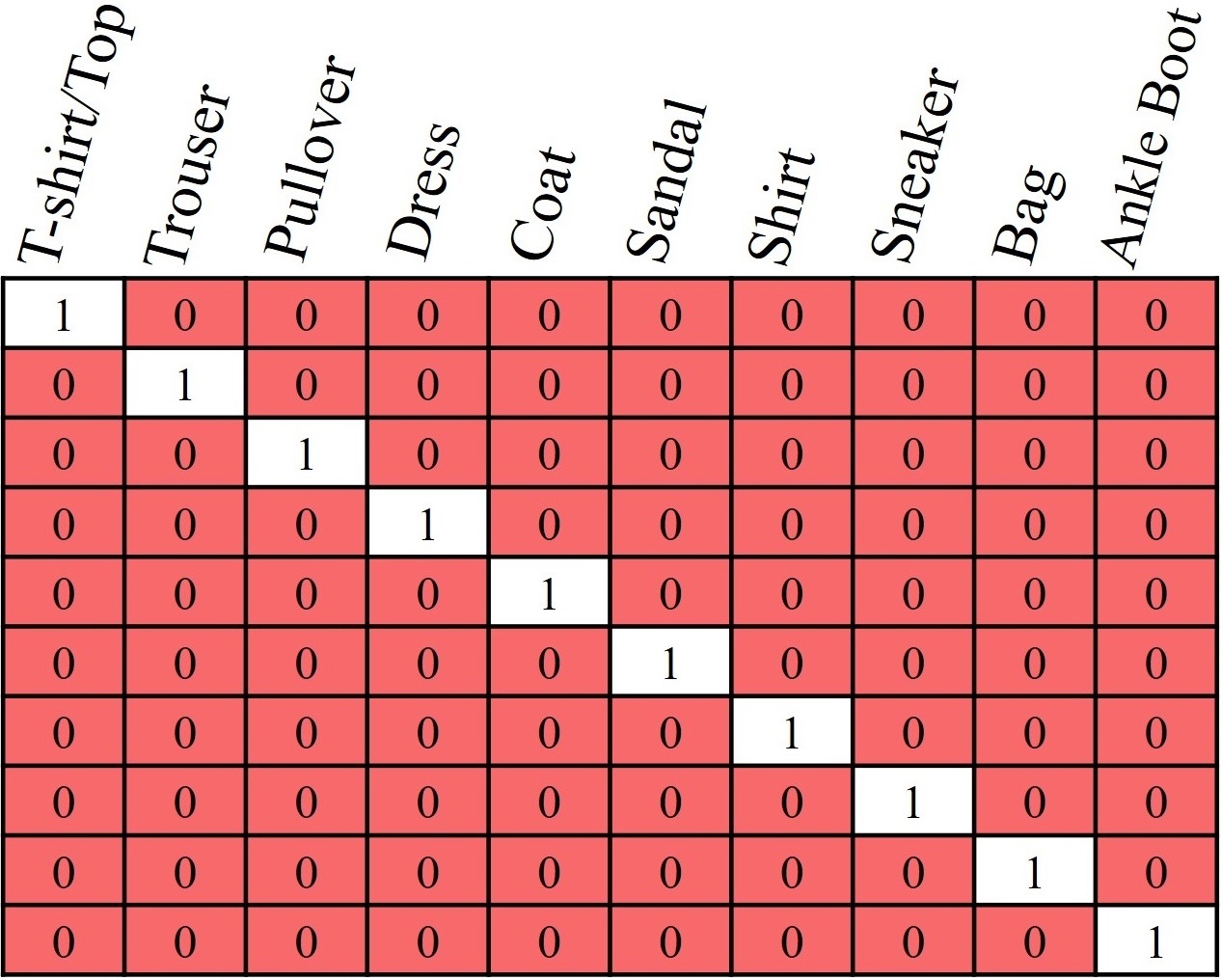}
         \caption{1-hot}
         \label{out_1hot}
     \end{subfigure}
     \hspace{0.2cm}
     \begin{subfigure}[b]{0.27\textwidth}
         \centering
         \includegraphics[width=\textwidth]{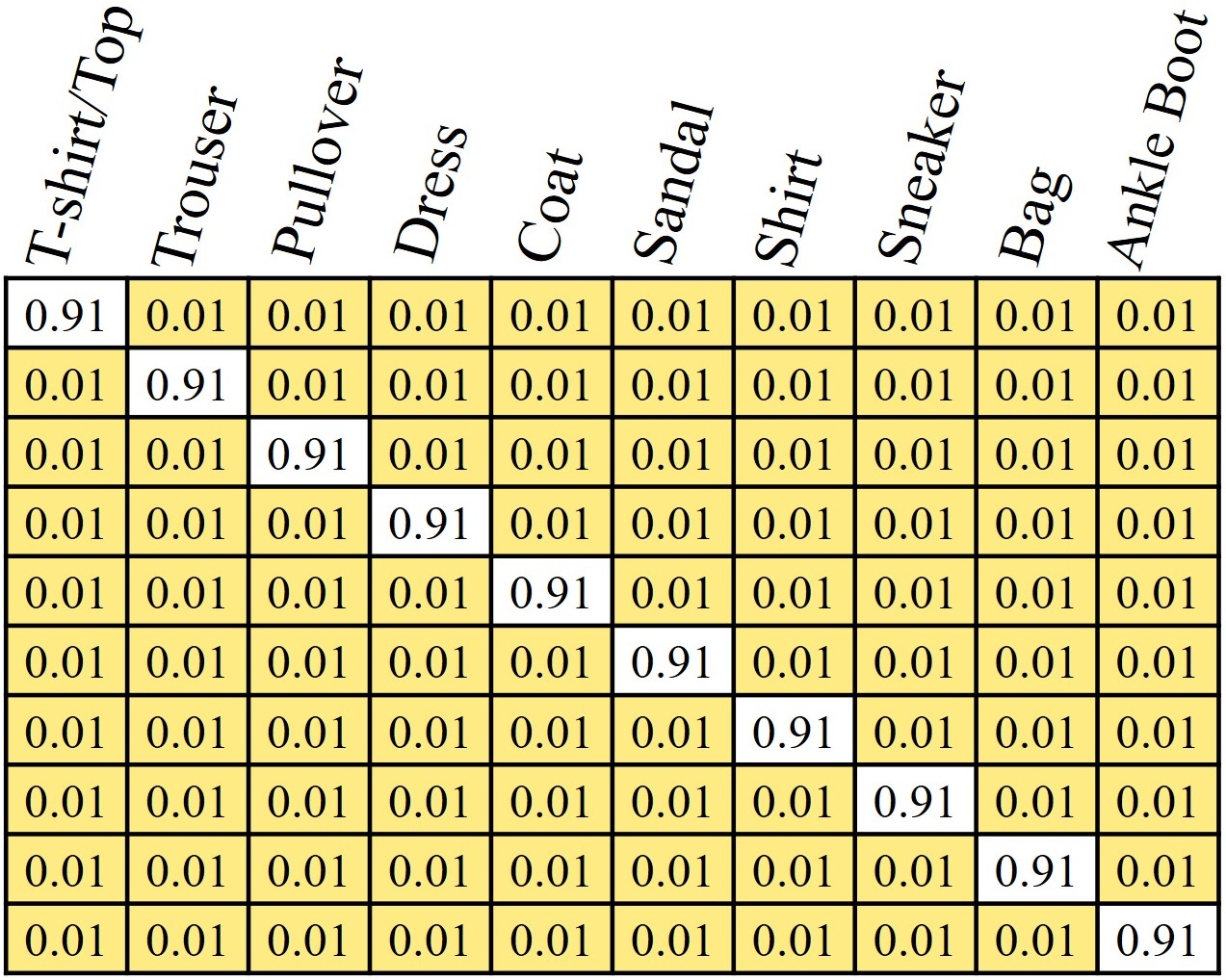}
         \caption{LS}
         \label{out_ls}
     \end{subfigure}
     \hspace{0.2cm}
     \begin{subfigure}[b]{0.27\textwidth}
         \centering
         \includegraphics[width=\textwidth]{figs/outp_ls.jpg}
         \caption{TFKD$_{reg}$}
         \label{out_tfkd}
     \end{subfigure}
     \begin{subfigure}[b]{0.27\textwidth}
         \centering
         \includegraphics[width=\textwidth]{figs/outp_1hot.jpg}
         \caption{OLS}
         \label{out_ols}
     \end{subfigure}
     \hspace{0.8cm}
     \begin{subfigure}[b]{0.27\textwidth}
         \centering
         \includegraphics[width=\textwidth]{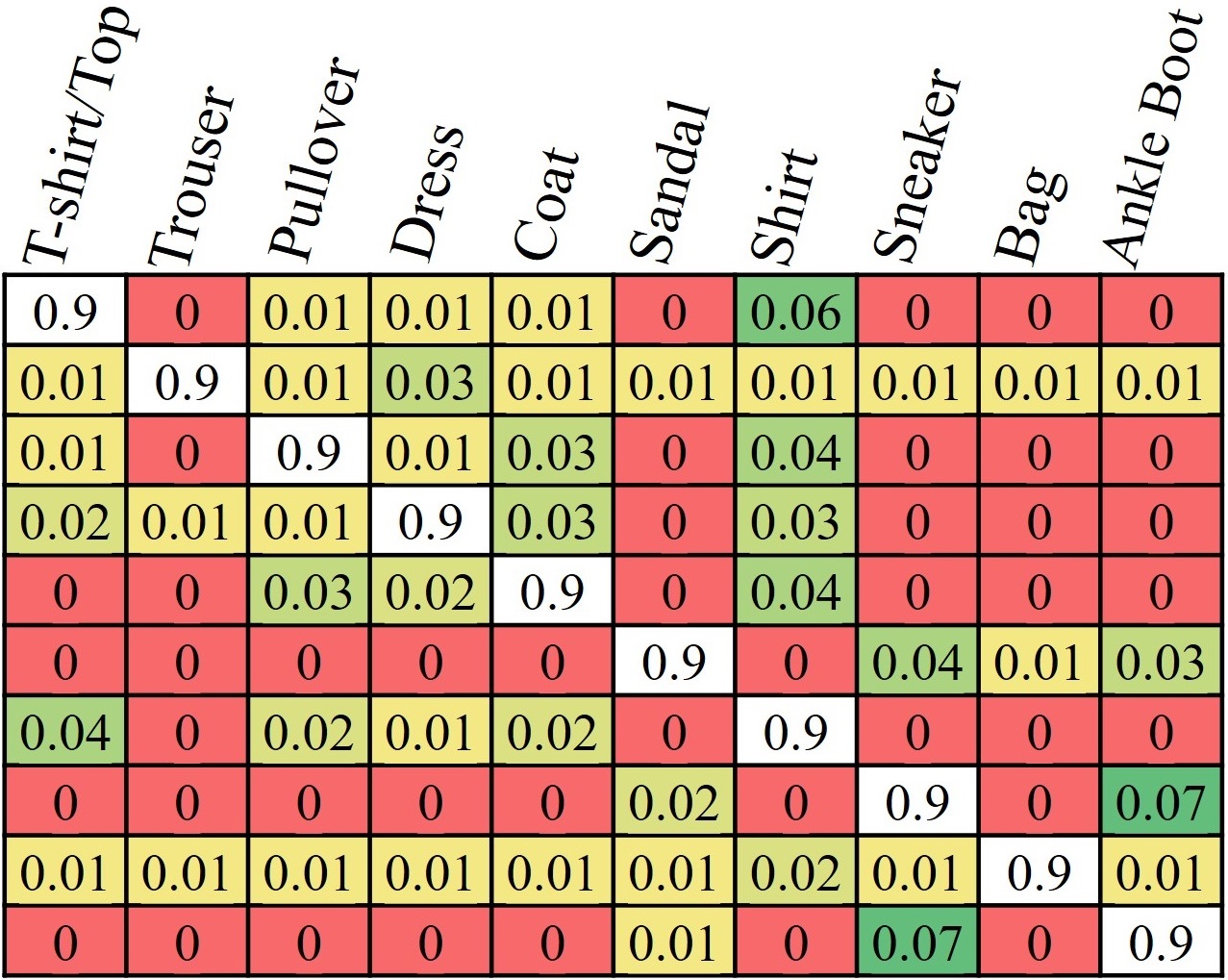}
         \caption{LS\texttt{++}}
         \label{out_lspp}
     \end{subfigure}
    \caption{Class-wise output probabilities on the training set of FashionMNIST dataset. }
    \label{fig:probs_out}
\end{figure*}

Label Smoothing$\texttt{++}$ consistently outperformed all compared approaches across different modalities, datasets, and networks. 
This underscores the superiority of learned mixing probability values over fixed or computed values. 
Furthermore, we showcase the impact of different methods on the final output probabilities on the FashionMNIST dataset in Figure \ref{fig:probs_out}.
Figure \ref{out_ols} exposes OLS collapsing to 1-hot training labels. Figure \ref{out_ls} and \ref{out_tfkd} demonstrate the disruption of inter-class relationships by uniform training labels. Conversely, Figure \ref{out_lspp} validates LS$\texttt{++}$, preserves inter-class relationships, and regularizes outputs.

% Figure \ref{out_ols} illustrates OLS collapsing to 1-hot training labels. In Figures \ref{out_tfkd} and \ref{out_ls} demonstrate employing a uniform distribution for generating regularized labels undermines interclass relations. Conversely, Figure \ref{out_lspp} validates Label Smoothing$\texttt{++}$ for preserving inter-class relations while regularizing outputs.

% In Figure \ref{out_ols}, it's evident that OLS collapses to 1-hot training labels. 
% Figures \ref{out_tfkd} and \ref{out_ls} demonstrate how utilizing a uniform distribution to generate regularized labels undermines interclass relationships. Finally, Figure \ref{out_lspp} validates the efficacy of Label Smoothing$\texttt{++}$ in both preserving interclass relations and regularization of outputs.

\begin{figure*}[!t]
\centering
     \begin{subfigure}[b]{0.32\textwidth}
         \centering
         \includegraphics[width=\textwidth]{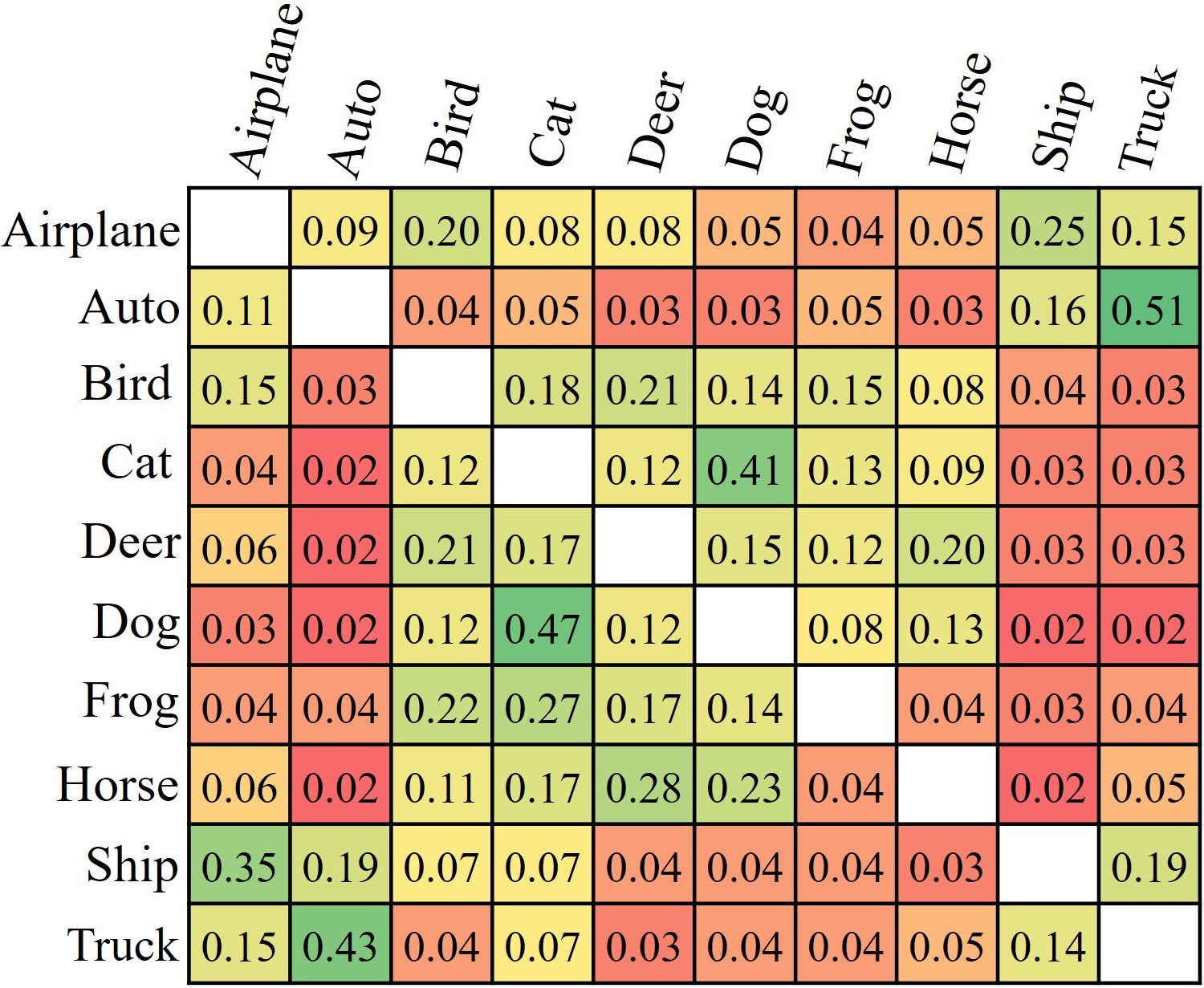}
         \caption{CIFAR-10}
     \end{subfigure}
     \hfill
     \begin{subfigure}[b]{0.32\textwidth}
         \centering
         \includegraphics[width=\textwidth]{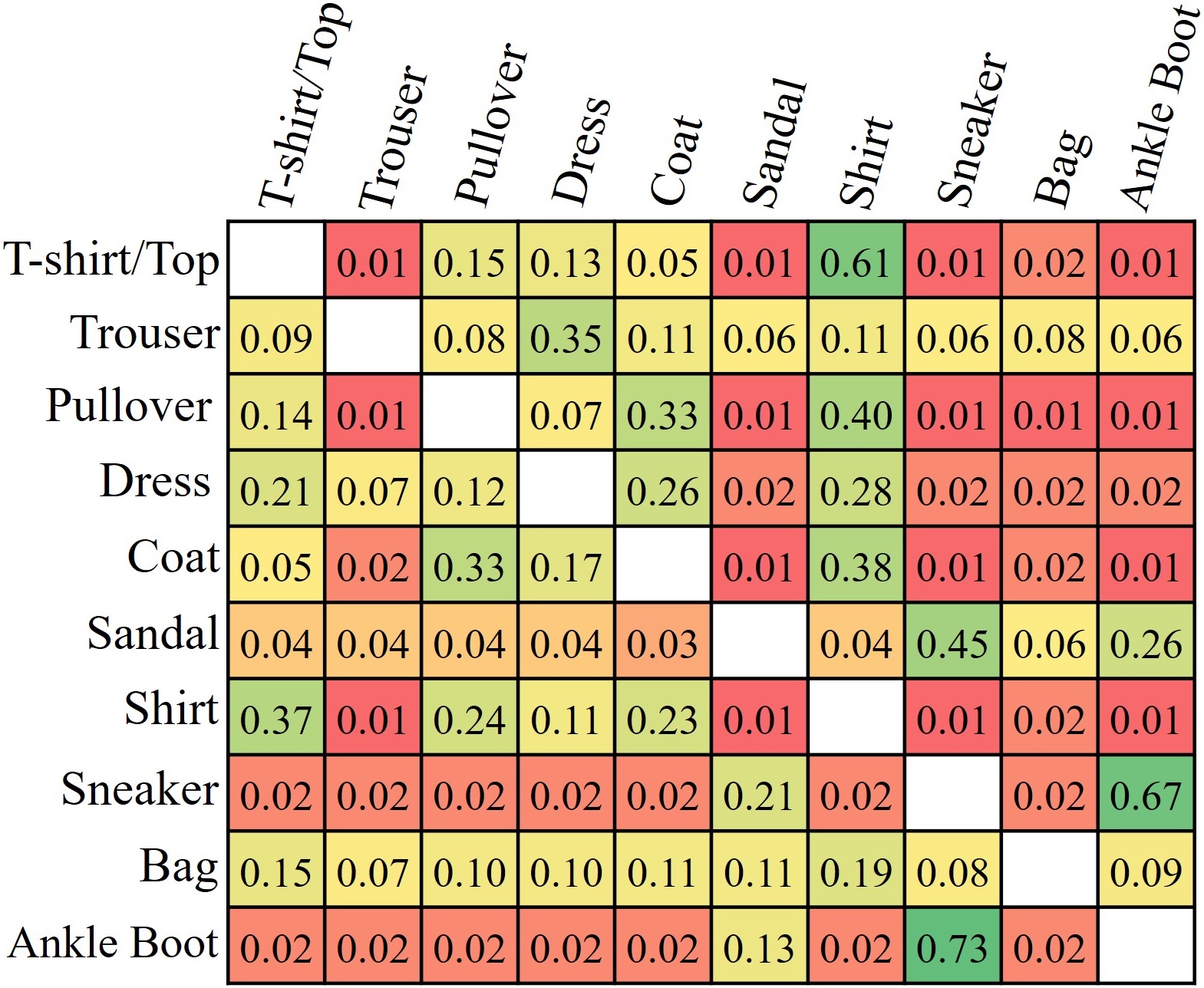}
         \caption{FashionMNIST}
     \end{subfigure}
     \hfill
     \begin{subfigure}[b]{0.32\textwidth}
         \centering
         \includegraphics[width=\textwidth]{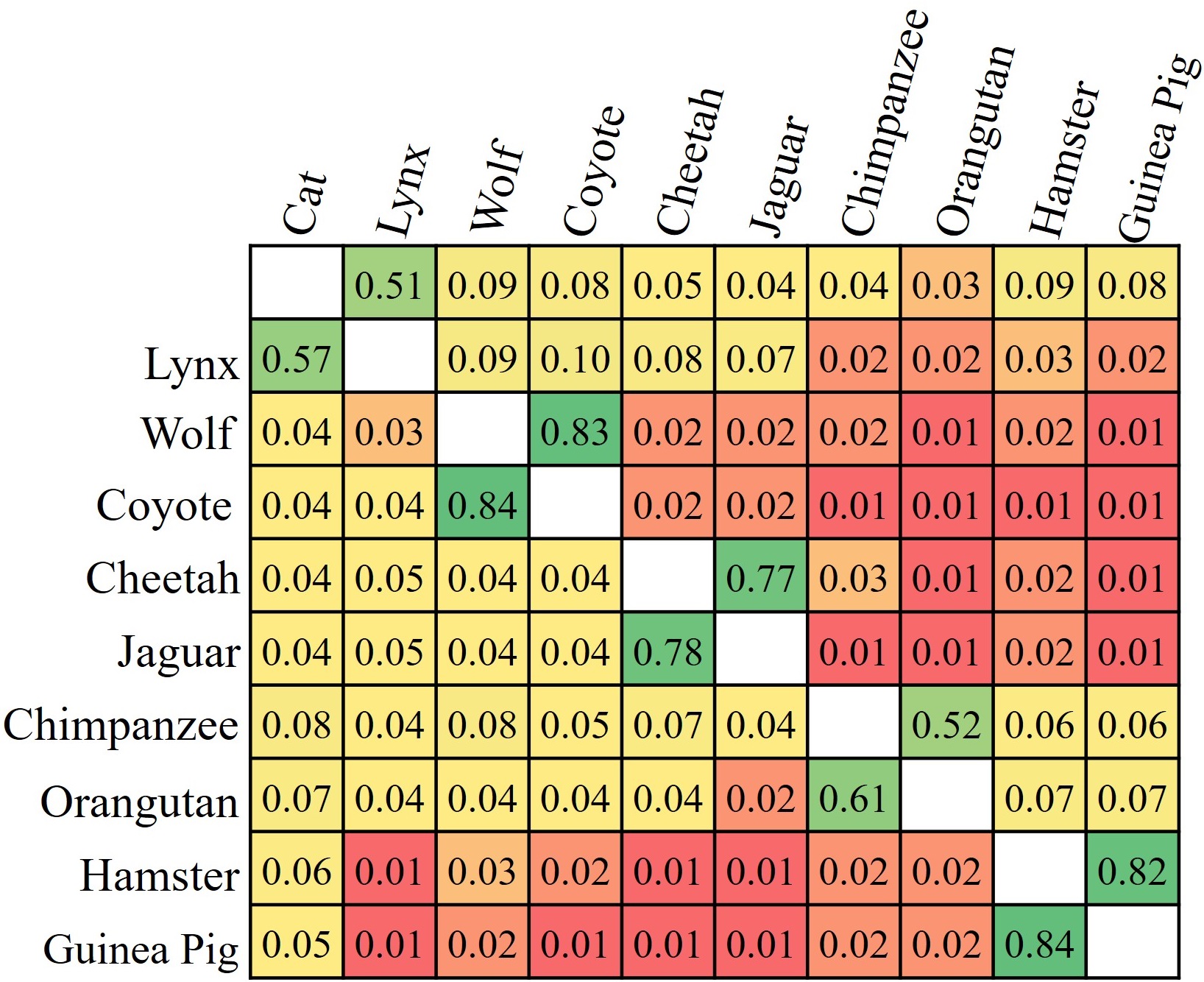}
         \caption{Animals-10N}
     \end{subfigure}
     % \begin{subfigure}[b]{0.21\textwidth}
     %     \centering
     %     \includegraphics[width=\textwidth]{figs/c_svhn.jpg}
     %     \caption{SVHN}
     % \end{subfigure}
     % \begin{subfigure}[b]{0.32\textwidth}
     %     \centering
     %     \includegraphics[width=\textwidth]{figs/cm_fer.jpg}
     %     \caption{FER2013}
     % \end{subfigure}    
     \caption{Learned C-Matrices. We can observe that the network favors the semantically close classes while distributing the probabilities and in turn, learns the inter-class relationships.}
    \label{fig:cmatrix}
\end{figure*}

% Figure \ref{out_ols} shows OLS collapses to 1-hot training labels. Figure \ref{out_tfkd} and \ref{out_ls} show the use of uniform distribution for generating regularized labels destroys the interclass relations. Figure \ref{out_lspp} confirms the preservation of inter-class relations and regularization of outputs Label Smoothing$\texttt{++}$.

We also present learned $C$-Matrices on CIFAR10, FashionMNIST, and Animals-10N datasets in Figure \ref{fig:cmatrix}. 
The analysis reveals the network's inclination towards assigning higher probabilities to semantically proximate classes. 
For instance, in the CIFAR10 dataset, the network exhibits a preference for classes like \textit{Cat} for \textit{Dog}, which are semantically closer. 
% Similarly, in FashionMNIST, the network favors classes such as \textit{Shirt}, and \textit{Pullover} for \textit{T-shirt}. 
Animals-10N is a fine-grain classification dataset and presents an interesting scenario with 5 pairs of confusing animals. 
The network consistently assigns probabilities to animals within each pair, considering them as the closest alternatives. 
% We did not find any impact of having a large number of classes on this behavior. 
% The network always learns the inter-class relationships among classes, even in the case of Tiny-ImageNet, which has 200 classes.
We also show $C$-Matrix for the case of a large number of classes (CIFAR100) in the supplementary material.

% Figure \ref{out_ls} and \ref{out_lspp} authenticate disruption and preservation of inter-class relations by Label Smoothing and Label Smoothing$\texttt{++}$, respectively.
% Figure \ref{fig:probs_out}(b) confirms that vanilla Label Smoothing disrupts inter-class relationships and Figure \ref{fig:probs_out}(c)  authenticates the preservation of inter-class relationships in the output probabilities by Label smoothing$\texttt{++}$.
% Notably, we can observe the preservation of inter-class relationships in the output probabilities of Label smoothing$\texttt{++}$. 
% Conversely, Figure \ref{fig:probs_out}(b) authenticates that vanilla Label Smoothing disrupts inter-class relationships.

\section{Analysis}

\subsection{Cluster Visualization}
\label{sec:tsne_analysis}
\begin{figure*}[t]
\centering
     \begin{subfigure}[b]{\textwidth}
         \centering
         \includegraphics[width=\textwidth]{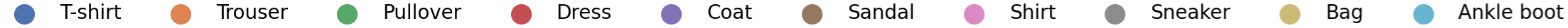}
     \end{subfigure}\\
     \begin{subfigure}[b]{0.25\textwidth}
         \centering
         \frame{\includegraphics[width=\textwidth]{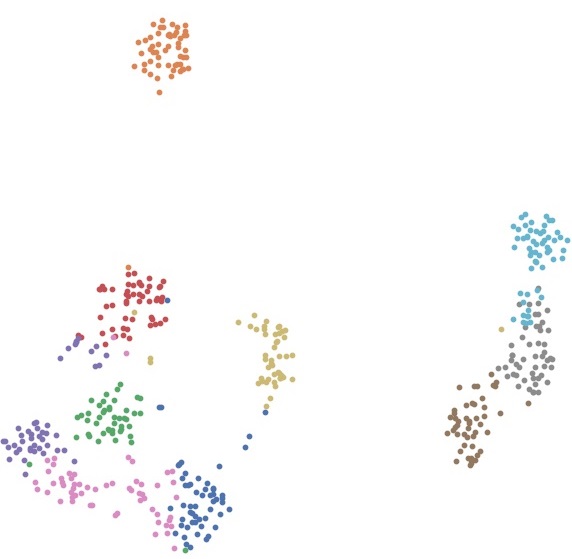}}
         % \caption{1-hot}
     \end{subfigure}
     \hspace{0.7cm}
     \begin{subfigure}[b]{0.25\textwidth}
         \centering
         \frame{\includegraphics[width=\textwidth]{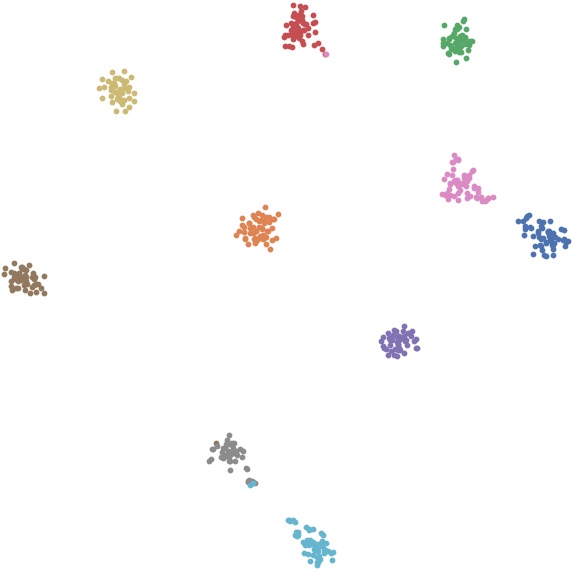}}
         % \caption{Label Smoothing}
     \end{subfigure}
     \hspace{0.7cm}
     \begin{subfigure}[b]{0.25\textwidth}
         \centering
         \frame{\includegraphics[width=\textwidth]{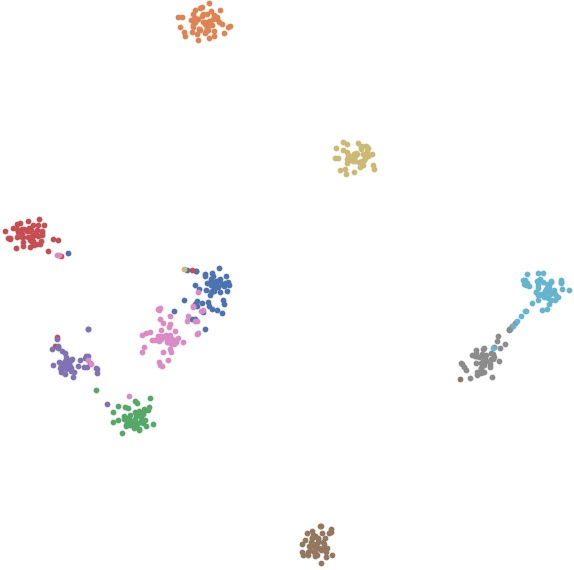}}
         % \caption{ls\texttt{+}\texttt{+}}
     \end{subfigure} \\[0.1cm]
          \begin{subfigure}[b]{0.28\textwidth}
         \centering
         \includegraphics[width=\textwidth]{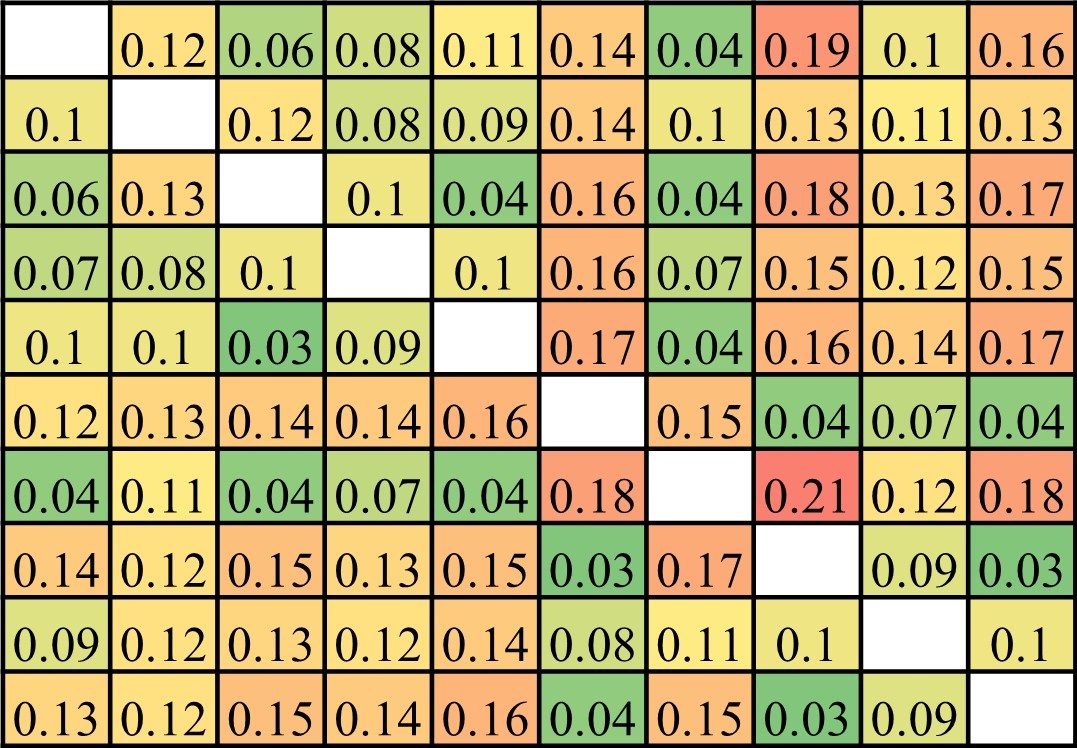}
         \caption{1-hot}
     \end{subfigure}
     \hspace{0.35cm}
     \begin{subfigure}[b]{0.28\textwidth}
         \centering
         \includegraphics[width=\textwidth]{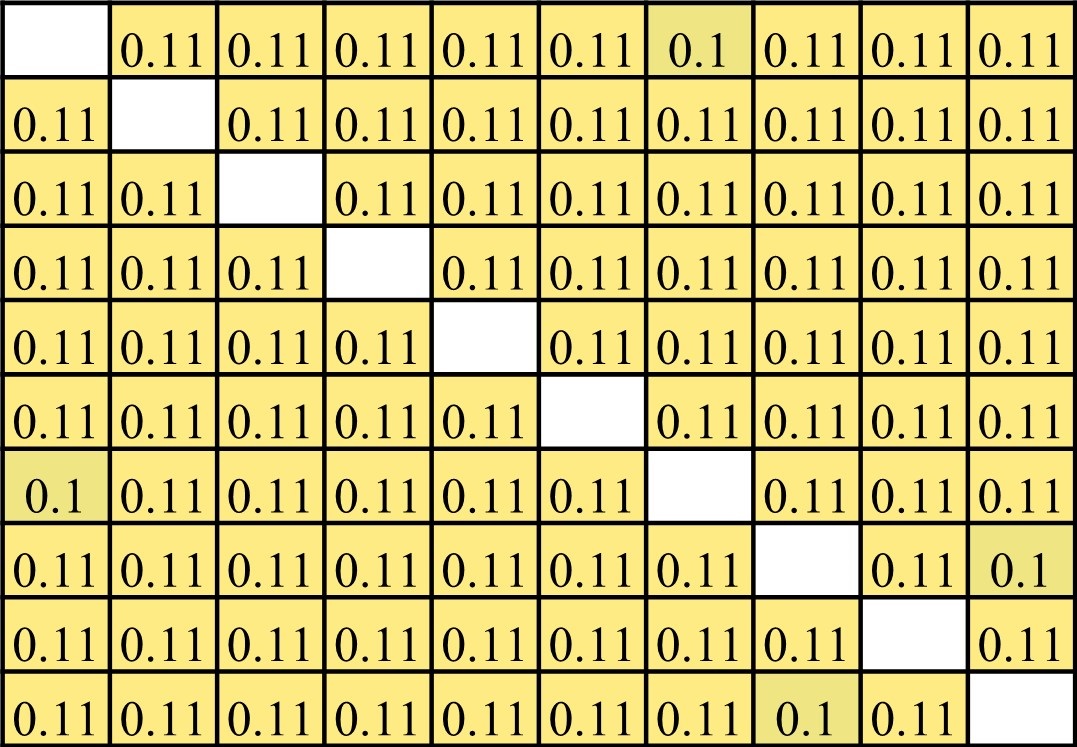}
         \caption{Label Smoothing}
     \end{subfigure}
     \hspace{0.35cm}
     \begin{subfigure}[b]{0.28\textwidth}
         \centering
         \includegraphics[width=\textwidth]{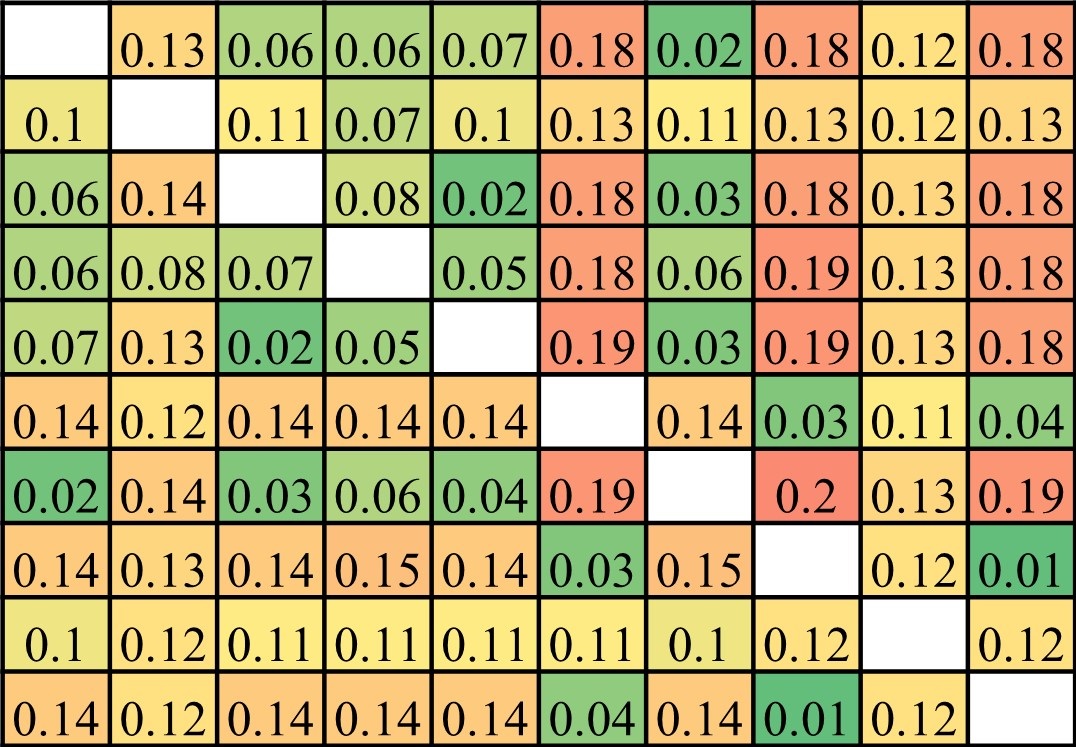}
         \caption{Label Smoothing$\texttt{++}$}
     \end{subfigure}
     \caption{In the upper row, we present TSNE visualizations of various approaches on the FashionMNIST dataset. 
     The lower row shows L$_1$-normalized cosine distance among the class cluster centers. 
     1-hot targets result in dispersed clusters, while Label Smoothing and Label Smoothing$\texttt{++}$ exhibit more compact clusters.
    However, Label smoothing places all clusters at an equal distance, effectively eliminating inter-class relationships. 
    In contrast, Label Smoothing$\texttt{++}$ maintains similar inter-class relationships as observed in 1-hot training.
    }
    \label{fig:tsne_with_dist}
\end{figure*}

In Figure \ref{fig:tsne_with_dist}, the top row showcases TSNE visualizations \cite{tsne} of FashionMNIST's training set. Notably, employing 1-hot targets results in dispersed clusters, while Label Smoothing and Label Smoothing$\texttt{++}$ yield more compact ones. Compact clusters are pivotal in minimizing collisions and enhancing generalization capabilities.
In the bottom row of Figure \ref{fig:tsne_with_dist}, we analyze the $L_1$-normalized cosine distance among class cluster centers. Label smoothing evenly spaces out all clusters, effectively eliminating inter-class relationships. 
On the other hand, Label Smoothing$\texttt{++}$ has the optimal effect that generates compact clusters, reduces overconfidence, and achieves high generalization while preserving inter-class relationships.

\begin{table}[t]
    \small
    \begin{center}
    % \begin{NiceTabular}{|p{1.5cm}|
    % >{\centering\arraybackslash}p{0.55cm} >{\centering\arraybackslash}p{0.7cm}|
    % >{\centering\arraybackslash}p{0.55cm} >{\centering\arraybackslash}p{0.7cm}|
    % >{\centering\arraybackslash}p{0.55cm} >{\centering\arraybackslash}p{0.7cm}|}
    \begin{NiceTabular}{|l|c|c|c|}
        \toprule
        % Dataset  & \multicolumn{2}{c|}{CIFAR10} & \multicolumn{2}{c|}{CIFAR100} & \multicolumn{2}{c|}{TinyImageNet}\\
        Method & CIFAR10 & CIFAR100  & TinyImageNet\\
        \midrule
        Vanilla Cross-Entropy & 80.45 & 78.95 & 64.93 \\
        Symmetric Cross-Entropy-Original & 79.92 & 78.27 & 64.39 \\
        Symmetric Cross-Entropy-Ours & \textbf{81.19} & \textbf{79.33} & \textbf{65.07} \\
        \bottomrule
    \end{NiceTabular}
    \end{center}
    \caption{Ablation Study on training loss for Label Smoothing$\texttt{++}$. CE: Cross-entropy, SCE-Original: Original symmetric cross-entropy that updates all parameters, SCE-Ours: Our symmetric cross-entropy loss that updates different parameters.}
    \label{tab:ablation}
\end{table}

\begin{figure*}[t]
\centering
     \begin{subfigure}[b]{0.27\textwidth}
         \centering
         \includegraphics[width=\textwidth]{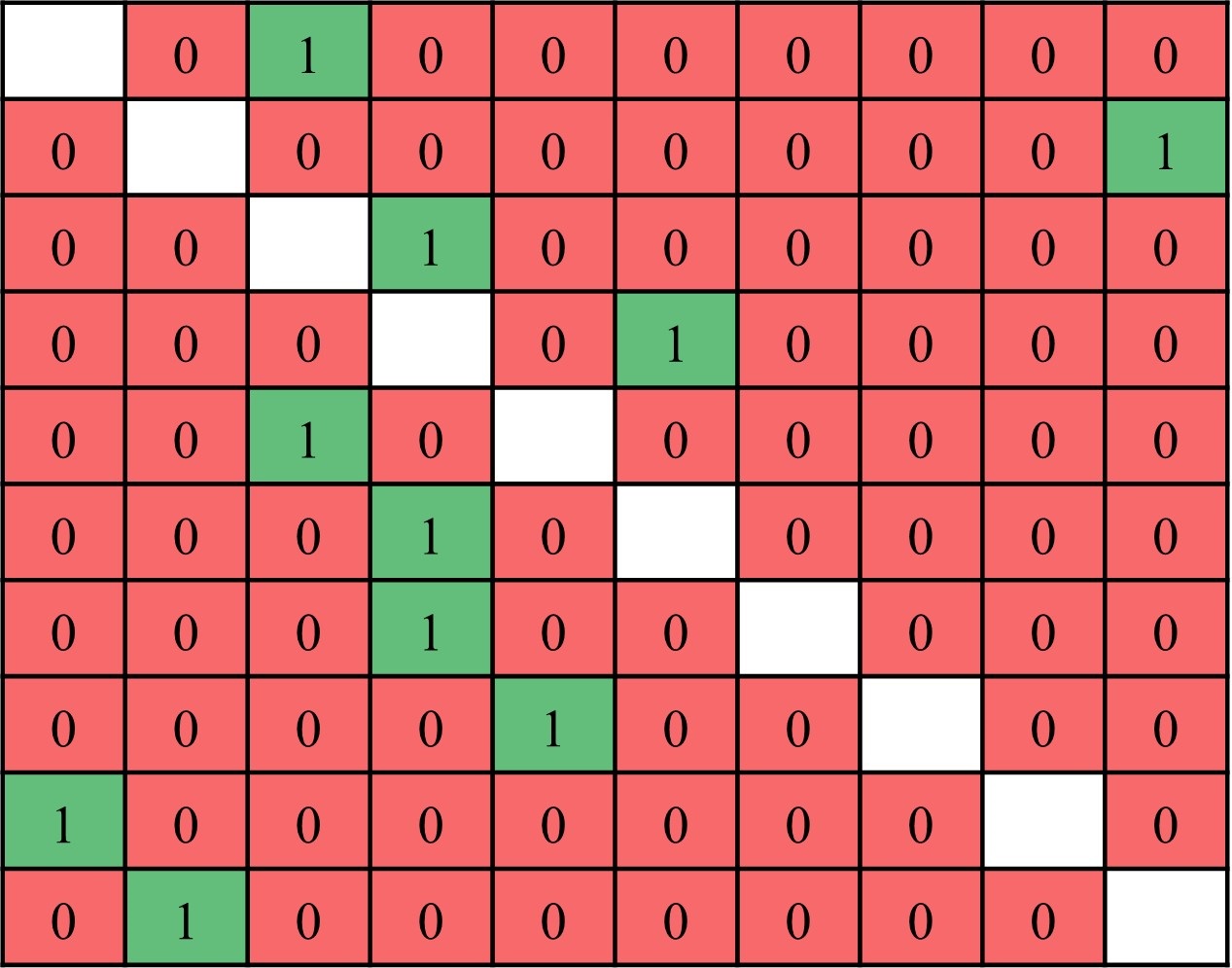}
         \caption{CE}
     \end{subfigure}
     \hspace{0.5cm}
     \begin{subfigure}[b]{0.27\textwidth}
         \centering
         \includegraphics[width=\textwidth]{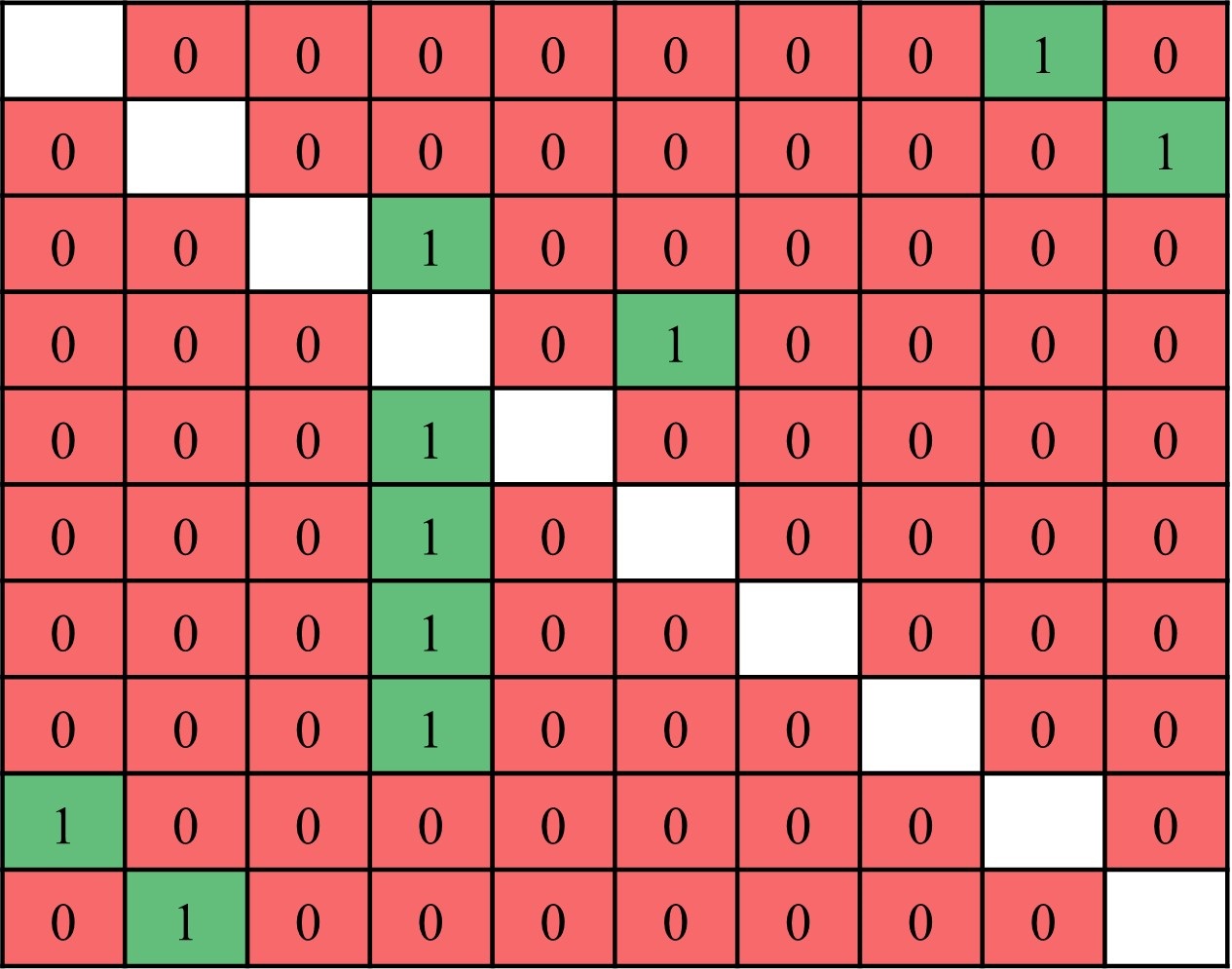}
         \caption{SCE-Original}
     \end{subfigure}
     \hspace{0.5cm}
     \begin{subfigure}[b]{0.27\textwidth}
         \centering
         \includegraphics[width=\textwidth]{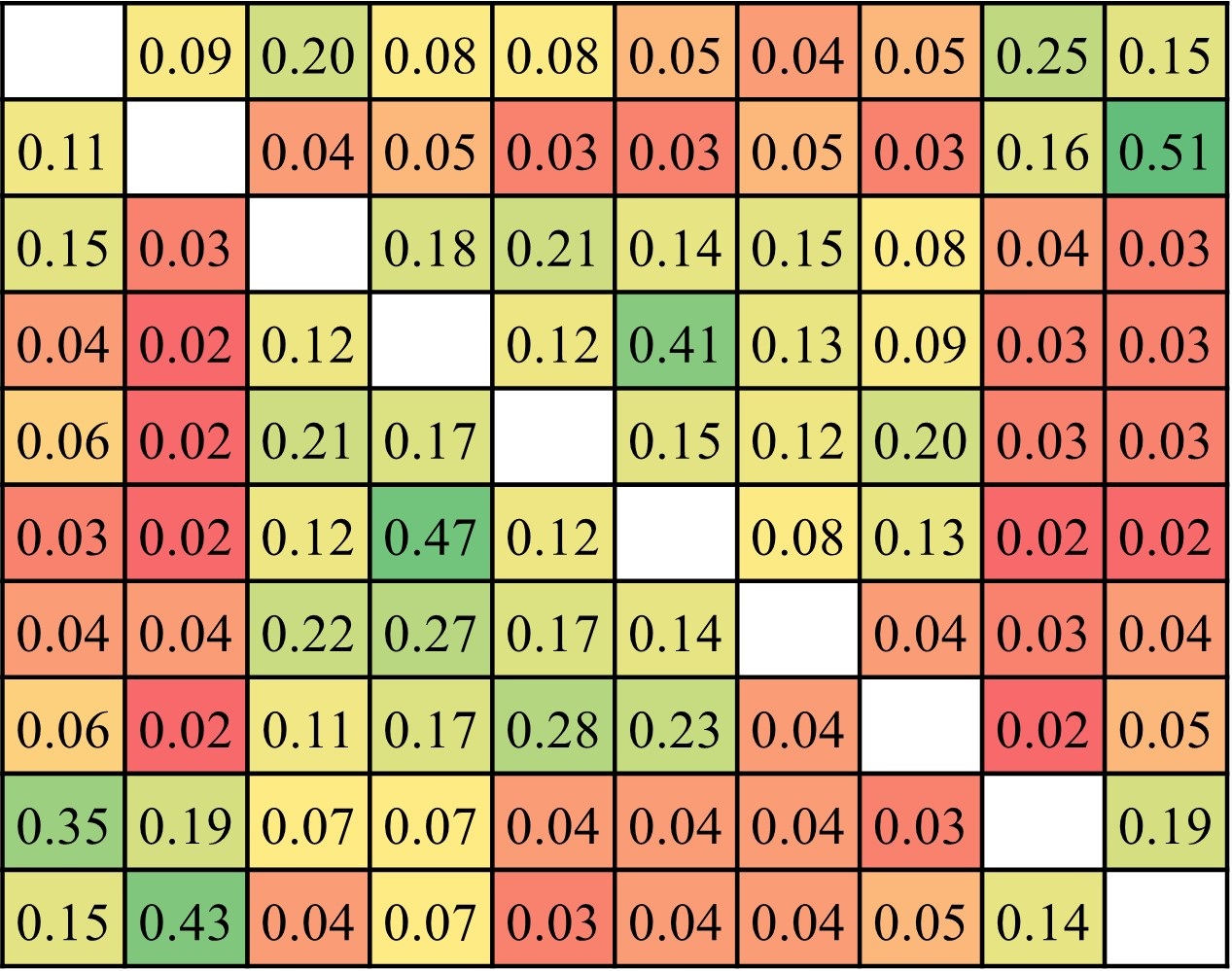}
         \caption{SCE-Ours}
     \end{subfigure}
    \caption{$C$-Matrix learned by cross-entropy, symmetric cross-entropy (SCE-original), and symmetric cross-entropy (SCE-ours) on CIFAR10. Cross-entropy and original symmetric cross-entropy result in a low entropy matrix and only our loss function provides the desired results. The targets can be generated by mixing these matrices with a 1-hot vector as per $\alpha$.}
        \label{fig:ablation_matrix}
\end{figure*}

\subsection{Ablation Study}
We perform an ablation study focusing on selecting training loss for Label Smoothing$\texttt{++}$. 
Our investigation highlights distinctions in outcomes and the acquired $C$ matrix through various loss functions, including standard cross-entropy (CE), original symmetric cross-entropy (SCE-Original), and our symmetric cross-entropy (SCE-Ours).
The matrix corresponding to CIFAR10 can be found in Figure \ref{fig:ablation_matrix}. 
Notably, both cross-entropy and original symmetric cross-entropy result in low entropy vectors. 
This outcome stems from the indirect entropy minimization loss, as elaborated in section \ref{sec:lspp}.
Contrastingly, our symmetric cross-entropy achieves the desired result by appropriately distributing probabilities.
The impact of this on generalization is illustrated in Table \ref{tab:ablation}, revealing a noticeable degradation in performance.

% In the upper row of Figure \ref{fig:tsne_with_dist}, we present TSNE \cite{tsne} visualizations of the training set of FashionMNIST. 
% Notably, 1-hot targets lead to dispersed clusters, while Label Smoothing, and LS$\texttt{++}$ exhibit compact clusters. 
% Compact clusters play a crucial role in minimizing collisions, representing a key factor contributing to their enhanced generalization capabilities.

% Moving to the second row of Figure \ref{fig:tsne_with_dist}, we examine the $L_1$-normalized cosine distance among the class cluster centers. 
% Label smoothing places all clusters at an equal distance, effectively eradicating inter-class relationships. 
% LS$\texttt{++}$ maintains similar inter-class relationships as 1-hot training. 
% LS$\texttt{++}$ has the optimal effect, it generates compact clusters, reduces overconfidence, and achieves high generalization while preserving inter-class relationships. 

% The distinction arises because LSP$_D$ employs a shared vector for generating targets across all classes, limiting flexibility in the arrangement of class clusters and consequently hindering inter-class relations.
% Breaking down the influences, LSP$_S$ has the least impact, primarily reducing overconfidence.
% LSP$_D$ exerts high constraints that create compact clusters, reduce overconfidence, and achieve high generalization but also disrupt inter-class relationships.
% LS$\texttt{++}$ has the optimal effect, it generates compact clusters, reduces overconfidence, and achieves high generalization while preserving inter-class relationships. 

\begin{table*}[t]
    \footnotesize
    % \small
    % \tiny
    \begin{center}
    \begin{NiceTabular}{|p{1.1cm}|
    >{\centering\arraybackslash}p{1.1cm} >{\centering\arraybackslash}p{1.1cm} 
    >{\centering\arraybackslash}p{1.1cm} >{\centering\arraybackslash}p{1.3cm} | 
    >{\centering\arraybackslash}p{1cm} >{\centering\arraybackslash}p{1.1cm} 
    >{\centering\arraybackslash}p{1.2cm} |}
    % \begin{NiceTabular}{|l|cccc|ccc|}
        \toprule
        Dataset & \multicolumn{4}{c}{CIFAR100} & \multicolumn{3}{c}{Tiny-ImageNet}\\
        \midrule
        Teacher  & R34$\rightarrow$R18 & R34$\rightarrow$R34 & R34$\rightarrow$R50 & R50$\rightarrow$SN & R50$\rightarrow$SN & R101$\rightarrow$SN & D121$\rightarrow$SN \\
        \midrule
        1-hot & 78.67 & 79.09 & 80.83 & 72.51 & 66.56 & 66.39 & 66.39\\
        LS & 79.40 & 80.15 & 81.15 & 71.70 & 66.58 & 66.74 & 66.80\\
        LS$\texttt{+}\texttt{+}$ & \textbf{79.90} & \textbf{80.38} & \textbf{81.16} & \textbf{72.64} & \textbf{66.89} & \textbf{67.00} & \textbf{67.43}\\
        \midrule
        PT-LS$\texttt{+}\texttt{+}$ & 79.75 & 79.90 & 80.93 & 72.59 & 64.10 & 64.76 & 64.75\\
        \bottomrule
    \end{NiceTabular}
    \end{center}
    \caption{Comparison of various teachers in the context of knowledge distillation, we consider three different teacher models: 1-hot, Label Smoothing (LS), and Label Smoothing++ (LS$\texttt{+}\texttt{+}$), each trained with their respective loss functions. Additionally, we introduce a proxy teacher model (PT-LS$\texttt{+}\texttt{+}$), where the $C$-Matrix learned by the teacher network (trained with LS$\texttt{+}\texttt{+}$) serves as the guiding information for training the student models.}
    \label{tab:kd_results}
\end{table*}

\subsection{Knowledge Distillation with a Proxy Teacher}
In knowledge distillation, a teacher network plays a guiding role in training a student network. 
The teacher network understands inter-class relations within a sample across all classes and generates regularized training labels for the student, enhancing its generalization. 
In our scenario, the $C$-Matrix serves as a proxy teacher (PT-LS$\texttt{+}\texttt{+}$) in the absence of a teacher network.
Leveraging the $C$-Matrix allows us to generate regularized training labels for each class instead of per sample. 
We conducted experiments on CIFAR-100 and Tiny-ImageNets for various transfer tasks. 
The student networks were trained using traditional cross-entropy loss, except that a teacher network provided training labels.
For the proxy teacher (PT-LS$\texttt{+}\texttt{+}$), training labels were created using the learned $C$-Matrix of the teacher network (trained with Label Smoothing$\texttt{+}\texttt{+}$). 

The results are in Table \ref{tab:kd_results}. As expected, knowledge distillation improves accuracy in all cases, with the network trained with LS$\texttt{+}\texttt{+}$ acting as the most effective teacher. 
The proxy teacher (PT-LS$\texttt{+}\texttt{+}$) achieves lower performance compared to other teachers but still outperforms training the network directly (refer Table \ref{tab:c100_results} and \ref{tab:ti_results}).
% The proxy teacher becomes valuable when a trained teacher model or sufficient memory is unavailable.
The biggest advantage of the proxy teacher is its independence from the teacher model's output, which can be computationally expensive. 
In our ResNet101 $\rightarrow$ ShuffleNet experiments on TinyImageNet, the proxy teacher took only half the time to train the student compared to traditional knowledge distillation.

\begin{table}[t]
    % \small
    \footnotesize
    \begin{center}
    \begin{NiceTabular}{|l|cccc|}
        \toprule
        Dataset  & CIFAR100 & FashionMNIST & TinyImageNet & ImageNet-100\\
        \midrule
        1-hot & 79.38 & 86.66 & 64.33 & 81.72\\
        \rowcolor{lightgray} LS$\texttt{++}$ & 80.25$_{\uparrow}$ & 87.47$_{\uparrow}$ & 65.07$_{\uparrow}$ & 82.22$_{\uparrow}$\\
        \midrule
        Cutout  & 80.11 & 88.36 & 65.86 & 82.86\\
        \rowcolor{lightgray}Cutout + LS$\texttt{++}$ & 80.44$_{\uparrow}$ & 88.92$_{\uparrow}$ & 66.53$_{\uparrow}$ & 83.04$_{\uparrow}$\\
        \midrule
        Mixup  & 81.31 & 88.48 & 66.17 & 81.88\\
        \rowcolor{lightgray}Mixup + LS$\texttt{++}$ & 81.46$_{\uparrow}$ &  88.59$_{\uparrow}$ & 66.41$_{\uparrow}$ & 82.88$_{\uparrow}$\\
        \midrule
        CutMix  & 81.95 & 88.11 & 68.50 & 83.50\\
        \rowcolor{lightgray}CutMix + LS$\texttt{++}$ & \textbf{82.24}$_{\uparrow}$ &  88.27$_{\uparrow}$ & \textbf{68.62}$_{\uparrow}$ & \textbf{83.66}$_{\uparrow}$\\
        \midrule
        RandAug  & 80.01& 92.40 & 65.87  & 82.88\\
        \rowcolor{lightgray}RandAug + LS$\texttt{++}$ & 80.24$_{\uparrow}$  & \textbf{92.85}$_{\uparrow}$ & 66.05$_{\uparrow}$ & 83.54$_{\uparrow}$\\
        \bottomrule
    \end{NiceTabular}
    \end{center}
    \caption{Application of Label Smoothing$\texttt{++}$ with Input Augmentations techniques - Co: Cutout, Mx: Mixup, Cx: CutMix, RA: RandAugment.}
    \label{tab:other_reg}
\end{table}

% \begin{table}[t]
%     % \small
%     \footnotesize
%     \begin{center}
%     \begin{NiceTabular}{|l|cccc|}
%         \toprule
%         Dataset  & CIFAR100 & FashionMNIST & TinyImageNet & ImageNet-100\\
%         \midrule
%         1-hot & 79.38 & 86.66 & 64.33 & 81.72\\
%         \rowcolor{lightgray} LS$\texttt{++}$ & 80.25$_{\uparrow}$ & 87.47$_{\uparrow}$ & 65.07$_{\uparrow}$ & 82.22$_{\uparrow}$\\
%         \midrule
%         Co \cite{cutout} & 80.11 & 88.36 & 65.86 & 82.86\\
%         \rowcolor{lightgray}Co + LS$\texttt{++}$ & 80.44$_{\uparrow}$ & 88.92$_{\uparrow}$ & 66.53$_{\uparrow}$ & 83.04$_{\uparrow}$\\
%         \midrule
%         Mx \cite{mixup} & 81.31 & 88.48 & 66.17 & 81.88\\
%         \rowcolor{lightgray}Mx + LS$\texttt{++}$ & 81.46$_{\uparrow}$ &  88.59$_{\uparrow}$ & 66.41$_{\uparrow}$ & 82.88$_{\uparrow}$\\
%         \midrule
%         Cx \cite{cutmix} & 81.95 & 88.11 & 68.50 & 83.50\\
%         \rowcolor{lightgray}Cx + LS$\texttt{++}$ & \textbf{82.24}$_{\uparrow}$ &  88.27$_{\uparrow}$ & \textbf{68.62}$_{\uparrow}$ & \textbf{83.66}$_{\uparrow}$\\
%         \midrule
%         RA \cite{randaug} & 80.01& 92.40 & 65.87  & 82.88\\
%         \rowcolor{lightgray}RA + LS$\texttt{++}$ & 80.24$_{\uparrow}$  & \textbf{92.85}$_{\uparrow}$ & 66.05$_{\uparrow}$ & 83.54$_{\uparrow}$\\
%         \bottomrule
%     \end{NiceTabular}
%     \end{center}
%     \caption{Application of Label Smoothing$\texttt{++}$ with Input Augmentations techniques - Co: Cutout, Mx: Mixup, Cx: CutMix, RA: RandAugment.}
%     \label{tab:other_reg}
% \end{table}

\subsection{Compatibility with Input Augmentations}
In this section, we assess the compatibility of Label Smoothing$\texttt{++}$ with input augmentation techniques such as Cutout, Mixup, Cutmix, and Randaugment. 
The results of this experiment are presented in Table \ref{tab:other_reg} using CIFAR100, FashionMNIST, Tiny-ImageNet, and ImageNet-100 datasets with ResNet34, ResNet18, ResNet18, and ResNet18, respectively.
Our findings indicate that label regularization seamlessly integrates with input regularization techniques. 
Employing input and label regularization together yields optimal performance, as evidenced by the results in the table.

\section{Conclusion}
In this paper, we introduced a label regularization technique termed Label Smoothing$\texttt{++}$, designed to enable neural networks to select their optimal training labels. 
Our approach uses different training labels for each class while ensuring that samples within the same class yield consistent outputs.
The training labels collectively form a $C$-Matrix which captures the inter-class relationships and serves as a proxy teacher for knowledge distillation.
Our proposed label regularization approach is compatible with input regularization and provides a performance boost when used together.
Extensive experimentation across various datasets demonstrates that Label Smoothing++ reduces overconfidence and promotes high generalization and inter-class relationships.\\

\noindent\textbf{Acknowledgements}
This work was supported in part by the following grants: National Institutes of Health Grant RF1AG073424,
National Institutes of Health Grant P30AG072980, and Arizona Department of Health Services Grant CTR057001.
Any opinions expressed in this material are those of the authors and do not necessarily reflect the views of funding agencies.

\bibliography{egbib}
\end{document}